\documentclass[11pt]{article}
\usepackage{nodalida2025}
\usepackage{hyperref}
\usepackage{times}
\usepackage{url}
\usepackage{latexsym}
\usepackage{graphicx}

\aclfinalcopy 

\title{Can summarization approximate simplification? A gold standard comparison}

\author{Giacomo Magnifico ~~~~~~ Eduard Barbu \\
  Institute of Computer Science \\
  University of Tartu \\
  Estonia \\
  {\tt \{giacomo.magnifico, eduard.barbu\}@ut.ee} \\
  }

\date{}

\begin{document}
\maketitle
\begin{abstract}
This study explores the overlap between text summarization and simplification outputs. While summarization evaluation methods are streamlined, simplification lacks cohesion, prompting the question: how closely can abstractive summarization resemble gold-standard simplification? We address this by applying two BART-based BRIO summarization methods to the Newsela corpus, comparing outputs with manually annotated simplifications and achieving a top ROUGE-L score of 0.654. This provides insight into where summarization and simplification outputs converge and differ.
\end{abstract}

\section{Introduction}

Text simplification can operate at various linguistic levels—semantic, syntactic, or lexical—using diverse strategies to achieve specific goals \cite{simplification-learners-2014, simplification-lowliteracy-2016, simplification-disability-2017, simplification-nlp-2021}. In practice, Automatic Text Simplification (ATS) transforms complex text into simpler versions by splitting sentences, shortening length, and simplifying vocabulary and grammar. The best English-language ATS models rely on parallel corpora like WikiSmall \cite{zhu-etal-2010-monolingual, zhang-lapata-2017-sentence}, aligning complex and simple sentences from standard and Simple English Wikipedias (originally 108,000 instances from 65,133 articles, currently 89,042).
The most valuable resource for text simplification is the Newsela corpus \newcite{newsela-xuetal-2015}, which includes 9,565 news articles professionally rewritten at multiple reading levels, with 1,913 original articles and four levels of simplification. However, it lacks the volume needed to train advanced deep-learning models effectively.

Simplification lacks standardized procedures and a common algorithm, partly due to the absence of a "native speaker of simplified language" \cite{siddhartan-2014}. The subjective nature of simplification also makes consistent methodology difficult \cite{simplification-survey-2022}. 
The evaluation metrics for simplification are similarly inconsistent. Some, like BLEU or Levenshtein distance \cite{levenshtein, bleu-papineni-2002}, focus on intrinsic grammatical features and struggle with semantic changes, while others, such as cosine distance, emphasize semantic similarity. By contrast, summarization metrics are well-established, even when imperfectly applied \cite{rogue-grusky-2023}. Furthermore, while the two tasks present some divergences in their focus (e.g. the relevance of information ordering, the choice of domain-agnostic lexicon, and the preference for short active forms instead of long passive forms), they remain convergent in producing shorter and poignant text. Given the state of things, we believe that comparing simplification with summarization could provide insights into their convergence.

This study investigates whether a state-of-the-art (SotA) summarization system can approximate manual simplification by comparing annotated simplifications with automated summarization. Starting with Newsela’s English documents, we process original articles with BRIO \cite{brio-liu-2022}, a SotA abstractive summarizer, applying document-wide and paragraph-by-paragraph summarization methods. We then evaluate each output set against the four simplification levels using ROUGE-L scores to measure similarity. Results indicate an average performance difference of 0.444, with paragraph-by-paragraph summarization achieving the highest score (0.654) at level 1, gradually decreasing through levels 2 to 4. While paragraph-by-paragraph summarization does not equate to manual simplification, it may serve as an effective preparatory step for manual annotators.

Background and related research are discussed in Section \ref{related-works}, with the experimental setup and findings detailed in Sections \ref{methods} and \ref{results}. A summary of the presented work, followed by the limits of the scope and suggestions for future research, are provided in Section \ref{discussion}.

\section{Related work}
\label{related-works}
The multifaceted nature of implementing text simplification has led to multiple works that share the goal of rewriting complex documents with simpler, more straightforward language. This is ultimately achieved by modifying the original text both lexically and syntactically as defined in \newcite{Truică2023}, either in an automated or manual way. Multiple works in the field have tackled different applications, from aiding people with disabilities \cite{simplification-disability-2013, simplification-disability-2017}, low-literacy adults \cite{simplification-lowliteracy-2009, simplification-lowliteracy-2016}, non-native learners \cite{simplification-learners-2009, simplification-learners-2014} to auxiliary systems to improve the effectiveness of other NLP tasks \cite{simplification-nlp-2013, simplification-nlp-2014, simplification-nlp-2016}. 

Due to the wide range of applications, a major subjectivity issue emerges when evaluating the different methods for simplification \cite{simplification-survey-2022}. Different scoring methods that have been utilized for simplification include: \textit{BLEU} \cite{bleu-papineni-2002}; \textit{TERp}, Translation Edit Rate plus, which computes the number of the three edit operations plus the inverse \cite{terp-snover-2009}; \textit{OOV}, Out Of Vocabulary, which measures the rate of oov words from a chosen simple vocabulary (e.g. Basic English list) \cite{oov-vuetal-2014}; \textit{changed}, measuring the percentage of the test examples
where the system suggested some change \cite{changed-hornetal-2014}; \textit{potential}, computing the proportion of instances in which at least one of the candidates generated is in the gold-standard \cite{simplification-lowliteracy-2016}; \textit{SARI}, the most recent, which performs a similar comparison to BLEU but is considered more reliable \cite{sari-xuetal-2016}.  

The general approach to text summarization is more streamlined, aiming to produce a shorter text than the input one while keeping all relevant information, defined as abstract or summary \cite{summ-moyadi-2016}. The most common approaches are naïve Bayes \cite{summ-bayes-kupiec-1995, summ-bayes-gambhir-2017}, swarm algorithms \cite{summ-swarm-jarraya-2012, summ-swarm-izakian-2015}, and sequence-to-sequence models \cite{summ-seq2seq-sutskever-2014, pegasus-zhang-2020}. 

A further distinction can be made between abstractive and extractive summarization methods \cite{summ-nazari-2019}. Where extractive methods produce text by concatenating selected parts of the original document, abstractive methods apply language generation techniques to produce a shorter document \cite{summ-jeek-2008, summ-gupta-2010}. Standard scoring methods for text summarisation are precision/recall measures and various instances of \textit{ROUGE} \cite{rouges-lin-2004, rogue-grusky-2023}, some examples being \textit{ROUGE-n, ROUGE-L,} and the most recent \textit{ROUGE-SEM} \cite{rougesem-zhang-2024}.

The Newsela corpus is a collection of 1,130 articles rewritten and simplified by professional editors, aimed at children of different grade levels \cite{newsela-xuetal-2015}. From each individual article, four different versions have been derived through the manual simplification process and labelled with a number from 1 to 4, representative of the level of simplification. Label 4 represents the most simplified output, suitable for a 3rd grader; label 3 represents an output suitable for a 4th grader; labels 2 and 1 identify outputs suitable for 6th and 7th graders. The original articles are suitable for 12th graders. 

Considering possible modifications to the dataset past the authors' presentation of their work, the corpus currently consists of 9,565 documents, of which 1,913 original articles.

\section{Experimental setup}
\label{methods}
For the purpose of this work, the architecture chosen to perform the summarization procedure was BRIO, a system presented in \newcite{brio-liu-2022} and based both on the BART architecture \cite{bart-lewisetal-2020} and the PEGASUS architecture \cite{pegasus-zhang-2020}. The choice was motivated by its state-of-the-art performance in summarization tasks, its ease of availability and implementation, and the double-model-based system that it employs. The dual nature of BRIO is the result of fine-tuning two different architectures on two different datasets with a specific training paradigm. Since the two datasets were characterized by longer texts \cite{cnndm-hermann-2015} and shorter texts \cite{xsum-narayanetal-2018}, the two backbones for the architecture keep these properties. Therefore, the BART-based BRIO was chosen as a summarizer for its performance with longer texts, as suggested by the original authors. 

\begin{figure*}[h]
    \centering
    \includegraphics[width=0.98\textwidth]{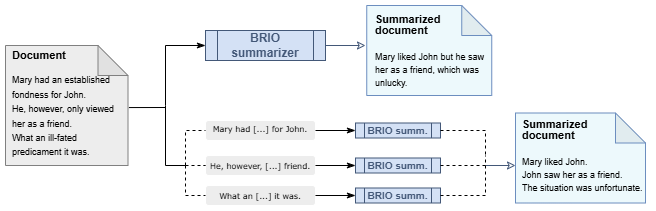}
    \caption{Representation of the processing pipeline for each article, showing the document-wide method (upper side) and paragraph-by-paragraph (lower side).}
    \label{fig:summarization}
\end{figure*}

The original articles from the Newsela corpus were then processed through the summarization model. For each article, two procedures were followed to produce different output documents: document-wide summarization and paragraph-by-paragraph summarization, as explained below. A graphic representation of the general procedure is provided in Figure \ref{fig:summarization}

\textbf{Document-wide.} The more intuitive application of text summarization, this method involved the generation of a single string containing the whole text by joining the various paragraphs and subsequently processing it with the summarizer model. Once the architecture produced an output string, it was written in a separate \texttt{*.txt} file.

\textbf{Paragraph-by-paragraph.} This summarization approach stems from the visual structure of academic texts, which usually separate topics and changes in content by dividing the document into paragraphs. Thus, the intuition was to make the architecture follow a similar pattern to preserve the content and produce a more effective summarization. This method implemented splitting the original text into paragraphs and processing each paragraph separately with the summarizer model. The resulting outputs were subsequently rejoined and written as a single document in a separate \texttt{*.txt} file.

Both procedures were applied to each of the original 1,913 English articles in the Newsela corpus, and the resulting two sets of summarized documents were compared to the simplified version produced by the editors. This was done by iterating through the different levels of simplification (1, 2, 3, and 4) and calculating the \textit{precision, recall} and \textit{ROUGE F1} score between each simplified version of the document and the summarized version of it. The resulting evaluation was stored, and the average was calculated level-wise for each metric with the scores from the whole set. Then, the scoring procedure was repeated for the remaining summarized set. The chosen evaluation score was \textit{ROUGE-L} as it was both a part of the original BRIO publication \cite{brio-liu-2022} and a statistic based on Long Common Sequence (LCS) \cite{lcs-linoch-2004}, which made it well suited to measure the grammatical integrity, keyword conservation and coherence in the summarized texts.

\section{Results}
\label{results}

\begin{figure*}[t]
    \centering
    \includegraphics[width=0.49\linewidth]{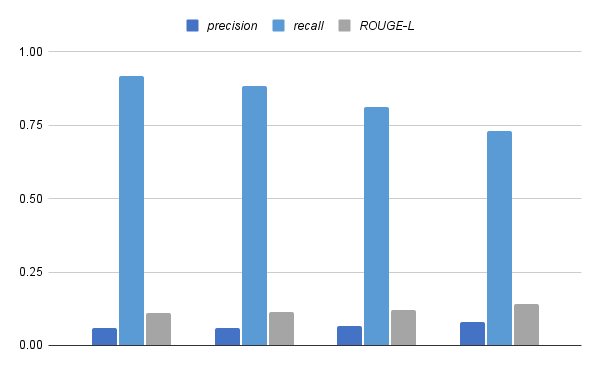}\hfill
    \includegraphics[width=0.5\linewidth]{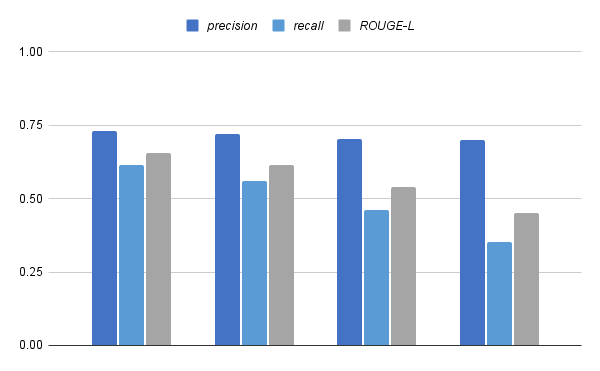}
    \caption{Comparison between the different levels of simplified text (1 to 4, left to right) and the two automated types of summarization. On the left is the performance of the document-wide summarization, on the right the performance of the paragraph-by-paragraph method.}
    \label{fig:scores}
\end{figure*}

The average scores for the three evaluation metrics used in comparing the human-produced simplification and the automated summarization are available in Table \ref{tab:scores}. To provide an easier analysis, the scores have been divided by the level of simplification taken under scrutiny and the type of summarization procedure performed on the original articles. The upper section of Table \ref{tab:scores} provides the average evaluation score between all the documents summarized with the first method mentioned in Section \ref{methods} and their simplified equivalent for each level. The second summarization method, paragraph-by-paragraph, is evaluated in the lower part of the Table.

\begin{table}[h]
    \centering
    \begin{tabular}{lccc}
        \textbf{Level} & \textbf{Precision} & \textbf{Recall} & \textbf{ROUGE-L} \\ \hline
        \multicolumn{4}{c}{DOCUMENT}   \\ \hline
        label 1     & 0.058     & 0.918  & 0.109   \\
        label 2     & 0.061     & 0.884  & 0.113   \\
        label 3     & 0.066     & 0.811  & 0.122   \\
        label 4     & 0.078     & 0.731  & 0.141   \\ \\ \hline
        \multicolumn{4}{c}{PARAGRAPH}  \\ \hline
        label 1     & 0.731     & 0.615  & 0.654   \\
        label 2     & 0.721     & 0.561  & 0.616   \\
        label 3     & 0.703     & 0.461  & 0.541   \\
        label 4     & 0.699     & 0.354  & 0.451  
    \end{tabular}

    \caption{Average precision, recall and ROUGE-L scores when comparing the summarization output against the different levels of manually simplified articles. The table is divided according to the two types of summarization techniques presented, \textit{document-level} and \textit{paragraph-level}.}
    \label{tab:scores}
\end{table}

To make the gap in scores and the variability in summarization performance through the different processes more apparent, two graphic representations of the average scores are provided in Figure \ref{fig:scores}. The data corresponds to the document-wide summarization method on the left side and the paragraph-by-paragraph method on the right. 

When comparing the results from the two processes, the overall difference in balance between precision/recall for the document-wide summarization method is immediately noticeable. Even considering the progressive improvement of the precision rate and the lowering of the recall score, the minimum gap between the two is 0.653. The first hypothesis was that it was due to the summarizer generating lengthy and repetitive summaries; however, a quick analysis of the outputs confirmed the variety in length and the production of shorter documents than their input. Therefore, the more plausible hypothesis is that while the longest common sentences between manual simplification and automated summarization are recalled in the text (most likely the keywords), the structural lexicon and syntactical choices of the simplified version would not appear through document-wide summarization. Consequently, this can lead to the poor similarity between the two document types and the convolution of information through summarization, a hypothesis corroborated by the low \textit{ROUGE-L} score.

On the other side of Figure \ref{fig:scores}, the scores provide a better-looking picture of the paragraph-by-paragraph performance. With a \textit{ROUGE-L} score of 0.566 averaged between all levels of simplification, the similarity between the simplified and summarized versions is noticeable. Although they perform better when compared to lower levels of simplification than to more simplified documents, the summarized outputs obtained through paragraph-by-paragraph processing perform well enough to justify further investigation and analysis. Our hypothesis for the better performance of the paragraph-by-paragraph, when compared to the document-wide processing, lies in the nature of the process: a block-by-block iteration might be more similar to the manually performed annotation than a text-wide transformation is.

Worth of notice for the production of these results was the difference in time requirements between the first summarization method and the second when operating on an average machine (16 GB RAM, 8 cores, 2,90 GHz CPU). The time elapsed for the paragraph-by-paragraph processing method was greatly increased, ranging between 10x and 50x more for each iteration and thus requiring several minutes instead of seconds. While the reason behind this issue requires more investigation, with the current implementation, performing such a method on a large-scale dataset without some optimization or access to a powerful machine is not recommended. 

\section{Conclusions, limitations and future work}
\label{discussion}

In this work, the similarities between simplified and summarized text have been analysed through the automated summarization of articles from the Newsela corpus, performed with two different methods and compared to four levels of professional manual simplification representative of diverse school grade levels. By examining the results obtained by a ROUGE-L scoring comparison between our output and the manual standard, it is shown that the proposed paragraph-by-paragraph method is superior to a document-wide approach, with the highest score being 0.654. Hence, it is possible to claim that while automated summarization does not produce text similar enough to simplified documents to justify its substitution, it still produces text similar enough to be used as a baseline to perform simplification on - instead of starting from the original text.

However, there are important limitations to the currently chosen metric. As ROUGE-L cannot measure semantic similarity between instances, all sequences that are semantically correct but lexically different would not compute as "similar". Since abstractive summarization could generate text that is lexically different from the simplification golden standard but still effectively simplified, further analysis with semantically relevant metrics should be conducted. In addition, future work in this direction should implement ulterior thorough analyses with more refined metrics, such as ROUGE-SEM or SARI, along with a comparison between manual simplification, automated summarization and automated simplification algorithms. In particular, the latter could shed some light on the intrinsic similarities between simplification and summarization and help further investigate the potential interdisciplinary approaches to the text simplification field of research.

Further investigation into optimization procedures to make the most-performing methods available for lower-end machines should also be conducted to allow for wider access to the tools and improved effectiveness of summarizers as a simplification helping tool.

\section{Acknowledgments}
This research has been supported by the EKTB55 project "Teksti lihtsustamine eesti keeles".

\bibliographystyle{acl_natbib}
\bibliography{nodalida2025}

\end{document}